\documentclass{article}

\usepackage{arxiv}

\usepackage[utf8]{inputenc} 
\usepackage[T1]{fontenc}    
\usepackage{hyperref}       
\usepackage{xcolor}
\usepackage{url}            
\usepackage{booktabs}       
\usepackage{caption}
\usepackage{amsfonts}       
\usepackage{amsmath}
\usepackage{nicefrac}       
\usepackage{microtype}      
\usepackage{algorithm}
\usepackage{algpseudocode}
\usepackage{graphicx}
\usepackage{lipsum}		
\usepackage{cite}
\usepackage{doi}

\title{The Effect of Sampling Temperature on Problem Solving \\ in Large Language Models}

\date{} 					

\author{
    \hypersetup{urlcolor=black}
	\href{https://matthewrenze.com}{Matthew Renze} \\
	Johns Hopkins University\\
    \hypersetup{urlcolor=black}
	\href{mailto:mrenze1@jhu.edu}{\texttt{mrenze1@jhu.edu}} \\
    \And
    \hypersetup{urlcolor=black}
    \href{https://ep.jhu.edu/faculty/erhan-guven/}{Erhan Guven}\\
	Johns Hopkins University\\
    \hypersetup{urlcolor=black}
	\href{mailto:eguven2@jhu.edu}{\texttt{eguven2@jhu.edu}} \\
}

\renewcommand{\headeright}{}
\renewcommand{\undertitle}{}
\renewcommand{\shorttitle}{The Effect of Sampling Temperature on Problem Solving in Large Language Models}

\hypersetup{
    colorlinks=true,
    linkcolor=blue,      
    urlcolor=blue,       
    citecolor=blue,      
    filecolor=magenta    
    pdftitle={The Effect of Sampling Temperature on Problem Solving in Large Language Models},
    pdfauthor={Matthew Renze, Erhan Guven},
    pdfkeywords={Large Language Model, LLM, Sampling Temperature},
}

\begin{document}

\maketitle

\begin{abstract}
In this research study, we empirically investigate the effect of sampling temperature on the performance of Large Language Models (LLMs) on various problem-solving tasks. We created a multiple-choice question-and-answer (MCQA) exam by randomly sampling problems from standard LLM benchmarks. Then, we used nine popular LLMs with five prompt-engineering techniques to solve the MCQA problems while increasing the sampling temperature from 0.0 to 1.6. Despite anecdotal reports to the contrary, our empirical results indicate that changes in temperature from 0.0 to 1.0 do not have a statistically significant impact on LLM performance for problem-solving tasks. In addition, these results appear to generalize across LLMs, prompt-engineering techniques, and problem domains. All code, data, and supplemental materials are available on GitHub at: \url{https://github.com/matthewrenze/jhu-llm-temperature}.
\end{abstract}

\section{Introduction}

\subsection{Background}

In recent years, Large Language Models (LLMs) have revolutionized the field of artificial intelligence. The availability of open-source LLMs and pay-per-use APIs has allowed engineers to incorporate LLMs in their AI systems. However, prompt engineering and hyperparameter tuning are required to work effectively with LLMs.

Prompt-engineering techniques help LLMs solve complex problems, avoid hallucinations, and provide more accurate responses. For example, we can use techniques like chain-of-thought, tree-of-thought, self-criticism, and self-consistency to improve LLM performance \cite{Mialon2023, White2023}.

In addition, several inference hyperparameters can be adjusted to modify the LLM’s output at runtime. For example, hyperparameters like sampling temperature, top-k sampling, repetition penalty, and maximum token length all affect the LLM's output and performance \cite{OpenAI_API_Reference, Touvron2023, Wang_Liu2023}.

Despite significant interest in LLMs and progress in LLM best practices, many open questions remain about optimal prompt-engineering techniques and inference hyperparameters for LLMs. To complicate matters, various local optima may exist for LLMs, prompt types, and problem domains \cite{Wang_Liu2023}.

The prompt-engineering community has an abundance of opinions and anecdotal evidence regarding optimal prompt-engineering techniques and inference hyperparameter settings. However, we currently lack systematic studies and empirical evidence to support many of these claims. 

As a result, this paper aims to address the open question of the optimal LLM sampling temperature for problem-solving tasks. In addition, we aim to provide a systematic study with empirical results to add to the growing body of knowledge used to create LLM and prompt-engineering best practices.

\subsection{Sampling Temperature}

Sampling temperature is a hyperparameter of an LLM used in a temperature-based sampling process. It controls the randomness of the model’s output at inference time \cite{Ackley1985, Hinton2015, Wang2020, Wang_Liu2023}.

During each step of an LLM’s decoding process, the LLM uses the previous tokens to choose the next output token. The final layer of the LLM uses a softmax function to convert raw scores (logits) into probabilities.

In greedy sampling, the model will always choose the most likely next token. However, for probabilistic sampling, the next token is selected from a probability distribution.

Temperature sampling is a modification to the softmax function, which adjusts the resulting probability mass functions. In this modified softmax function, $v_k$ is the $k$-th vocabulary token, $l_k$ is the token’s logit, and $\tau$ is a constant temperature. See equation \ref{eq:temperature_sampling}.

\begin{equation}
\Pr(v_k) = \frac{e^{l_k / \tau}}{\sum_i e^{l_i / \tau}}
\label{eq:temperature_sampling}
\end{equation}

A lower temperature makes the output of the LLM more deterministic, thus favoring the most likely predictions. This conservativeness is captured by the model’s tendency to produce more repetitive, focused, and less diverse output based on the patterns most commonly seen in the training data \cite{Hinton2015, Wang2020, Wang_Liu2023}.

A higher temperature increases the randomness of the output, thus favoring more “creative” predictions. This creativity is captured by the model's willingness to explore more unconventional and less likely outputs. Higher temperatures can lead to novel text, diverse ideas, and creative solutions to problems \cite{Hinton2015, Wang2020, Wang_Liu2023}.

In the context of problem-solving, temperature can be seen as a trade-off between exploring and exploiting possible solutions within the solution space. Lower temperatures tend to exploit more probable solutions; higher temperatures explore the solution space more broadly.

\subsection{Choosing a Sampling Temperature}

Within the prompt-engineering community, there are a variety of opinions and best practices regarding the ideal sampling temperature for various problem-solving tasks \cite{Microsoft_Azure_OpenAI, Shieh_Prompt_Engineering}. 

Low sampling temperatures are recommended for tasks requiring precision and factual accuracy, such as technical writing, code generation, or question-answering \cite{Xu2022, Zhu2023}. However, higher temperatures are recommended for tasks requiring creativity, such as writing poetry, creating stories, or brainstorming.

Higher temperatures also increase the probability of model hallucination. Hallucination is a phenomenon where an LLM produces statistically probable responses that are factually incorrect or nonsensical. As a result, optimal temperature selection is also a balance between creativity and hallucination \cite{Lee2023}.

Practical guidelines for choosing a sampling temperature for a specific task or problem domain are often vague or anecdotal. Prompt-engineering guides often provide hypothetical examples of optimal sampling temperatures for various tasks. However, they rarely cite any sources or provide empirical evidence.\footnote{A few empirical studies exist that indicate sampling temperature does have an effect on LLM performance on some types of problem-solving tasks (e.g., code generation, engineering exams, etc.) \cite{Xu2022, Pursnani2023, Zhu2023}.}

As a result, the current state of choosing the optimal sampling temperature for specific problems is largely based on guesswork, gut instinct, non-systematic experimentation, and iterative refinement.\footnote{For example, OpenAI's GPT-3.5 API allowed users to set the sampling temperature from 0.0 to 1.0 with a default of 0.7. GPT-4's API expanded this range from 0.0 to 2.0 with a default of 1.0. No explanation from OpenAI has been provided for these default values or their change from GPT-3.5 to GPT-4 \cite{OpenAI_Forum}.}$^{,}$\footnote{Even the GPT-4 Technical Report explains that the authors used their “best-guess” when choosing sampling temperatures while evaluating GPT-4 on various benchmarks. See Appendix A in the GPT-4 Technical Report \cite{OpenAI_GPT4_Report}.}

\section{Methods}

\subsection{Models}

The models used in this research project comprise nine widely-used foundational LLMs. To complement our analysis, we also conducted experiments using five prompts created using commonly used prompt-engineering techniques.

First, we reviewed the prior literature to identify candidate LLMs commonly used for problem-solving tasks. We limited our candidate models to those that allowed the model's sampling temperature to be specified via their API \cite{OpenAI_ChatGPT, OpenAI_GPT4_Blog, OpenAI_GPT4_Report, Touvron2023}. See Table \ref{tab:models} for a list of LLMs used in the experiment.

\begin{table*}[ht]
    \begin{center}
        \begin{small}
            \begin{tabular}{lllrl}
                \toprule
                \textbf{Name} & \textbf{Vendor} & \textbf{Released} & \textbf{License} & \textbf{Source} \\
                \midrule
                Claude 3 Opus & Anthropic & 2024-03-04 & Closed & \cite{Anthropic2024A, Anthropic2024B} \\
                Command R+ & Cohere & 2024-04-04 & Open & \cite{Cohere2024A, Cohere2024B} \\
                Gemini 1.0 Pro & Google & 2023-12-06 & Closed & \cite{Pichai2023, GeminiTeam2023} \\
                Gemini 1.5 Pro (Preview) & Google & 2024-02-15 & Closed & \cite{Pichai2024, GeminiTeam2024} \\
                GPT-3.5 Turbo & OpenAI & 2022-11-30 & Closed & \cite{OpenAI_ChatGPT, OpenAIGPT35} \\
                GPT-4 & OpenAI & 2023-03-14 & Closed & \cite{OpenAI_GPT4_Blog, OpenAI_GPT4_Report} \\
                Llama 2 7B Chat & Meta & 2023-07-18 & Open & \cite{Meta2023, Touvron2023} \\
                Llama 2 70B Chat & Meta & 2023-07-18 & Open & \cite{Meta2023, Touvron2023} \\
                Mistral Large & Mistral AI & 2024-02-26 & Closed & \cite{MistralAI2024} \\                
                \bottomrule
            \end{tabular}
        \end{small}
    \end{center}
    \caption{LLMs used in the experiment.}
    \label{tab:models}
    \vskip -0.1in
\end{table*}

Next, we reviewed the existing literature for commonly used prompt-engineering techniques. We limited our candidate prompts to those that could be performed in a single request-and-response cycle with one-shot in-context learning. We excluded multi-step agents, few-shot learning, and model fine-tuning.

As a result, we selected five prompt-engineering techniques to construct our system prompts:
\begin{itemize}    
    \item \textbf{Baseline} - no prompt engineering; the LLM is instructed to return only a single multiple-choice answer as its output (e.g., ‘Answer(“C”)’ ).
    \item \textbf{Domain Expertise} – the system prompt specifies that the LLM is an expert in the problem domain of the exam (e.g., “medicine”) or the topic of the problem (e.g., “anatomy”)  \cite{White2023}.
    \item \textbf{Self-recitation} – the system prompt instructs the LLM to recite its own internal knowledge about the problem before answering the question \cite{Sun2023, White2023}.
    \item \textbf{Chain-of-Thought (CoT)} – the system prompt instructs the LLM to “think step-by-step” to encourage it to reason through the problem procedurally \cite{Kojima2022, Wei2022}.
    \item \textbf{Composite} – the system prompt combines domain expertise, self-recitation, chain-of-thought, and adds self-criticism \cite{Huo2023, Wang_Wang2023}. 
\end{itemize}

Finally, we provided the LLM with a single example problem-and-solution pair for one-shot in-context learning. The example solution was adapted for each prompt based on the prompt-engineering technique used. For example, the CoT prompt included a chain of thought in its solution. See Figure \ref{fig:one-shot-example} in the Appendix for a sample prompt.

\subsection{Data}

The test dataset used in this research study consists of a series of Multiple-Choice Question-and-Answer (MCQA) exams derived from widely used LLM performance benchmarks.

First, we reviewed the prior literature to identify benchmarks frequently used to evaluate LLMs. We limited our candidate benchmarks to those containing MCQA problems so that we could use correct-answer accuracy as our primary performance metric.

Next, we selected a set of problems that covered a range of problem domains (e.g., math, science, law, etc.) and difficulty levels (e.g., secondary school, university, etc.) These problem sets can be seen in Table \ref{tab:exams}.

\begin{table*}[ht]
    \begin{center}
        \begin{small}
            \begin{tabular}{lllrrl}
                \toprule
                \textbf{Problem Set} & \textbf{Benchmark} & \textbf{Domain} & \textbf{Questions} & \textbf{License} & \textbf{Source} \\
                \midrule
                ARC Challenge Test & ARC & Science & 1,173 & CC BY-SA & \cite{Clark2018} \\
                AQUA-RAT & AGI Eval & Math & 254 & Apache v2.0 & \cite{Zhong2023} \\
                Hellaswag Val & Hellaswag & Common Sense Reasoning & 10,042 & MIT & \cite{Zellers2019} \\
                LogiQA (English) & AGI Eval & Logic & 651 & GitHub & \cite{Zhong2023, Liu2020} \\
                LSAT-AR & AGI Eval & Law (Analytic Reasoning) & 230 & MIT & \cite{Zhong2023, Wang2021} \\
                LSAT-LR & AGI Eval & Law (Logical Reasoning) & 510 & MIT & \cite{Zhong2023, Wang2021} \\
                LSAT-RC & AGI Eval & Law (Reading Comprehension) & 260 & MIT & \cite{Zhong2023, Wang2021} \\
                MedMCQA Valid & MedMCQA & Medicine & 6,150 & MIT & \cite{Pal2022} \\
                SAT-English & AGI Eval & English & 206 & MIT & \cite{Zhong2023} \\
                SAT-Math & AGI Eval & Math & 220 & MIT & \cite{Zhong2023} \\
                \bottomrule
            \end{tabular}
        \end{small}
    \end{center}
    \caption{Problem sets used to create the multi-domain MCQA exam.}
    \label{tab:exams}
    \vskip 0.1in
    {\footnotesize Note: The GitHub repository for LogiQA does not include a license file. However, both the paper and readme.md file states that "The dataset is freely available."}
\end{table*}

Then, we converted the benchmark problems from their original data format into a standardized data structure using the JSON Lines (JSON-L) format \cite{Ward_JSON}. Our standardized set of exams allowed us to use the exams interchangeably without modifying the code in the test harness. See Figure \ref{fig:mcqa-sample} in the Appendix for a sample of an MCQA problem.

Finally, we created two MCQA exams of different sizes. We created a large exam with 1,000 questions by randomly sampling 100 problems from each of the ten problem sets. This 1,000-question (large) exam was used with GPT-3.5 to perform a detailed analysis of temperature across problem domains.

Additionally, we created a smaller exam of 100 questions by randomly sampling ten questions from each of the ten domain-specific problem sets. This 100-question (small) exam was used for our high-level analysis of sampling temperature across all nine models, all five prompt-engineering techniques, and extended temperature range (0.0-1.6).\footnote{We used the smaller 100-question exam due to cost and runtime considerations.}

\subsection{Process}

Our experiment was designed to test the problem-solving performance of LLMs across ten models, five prompt-engineering techniques, ten problem domains, 100 problems within each problem domain, and all viable sampling temperatures. For each combination of model, prompt, exam, and temperature, we instructed the LLM to answer each question ten times so we could assess the average correct-answer accuracy.

\begin{figure}[h]
    \begin{minipage}{0.48\textwidth}   
        \centering
        \includegraphics[width=\linewidth]{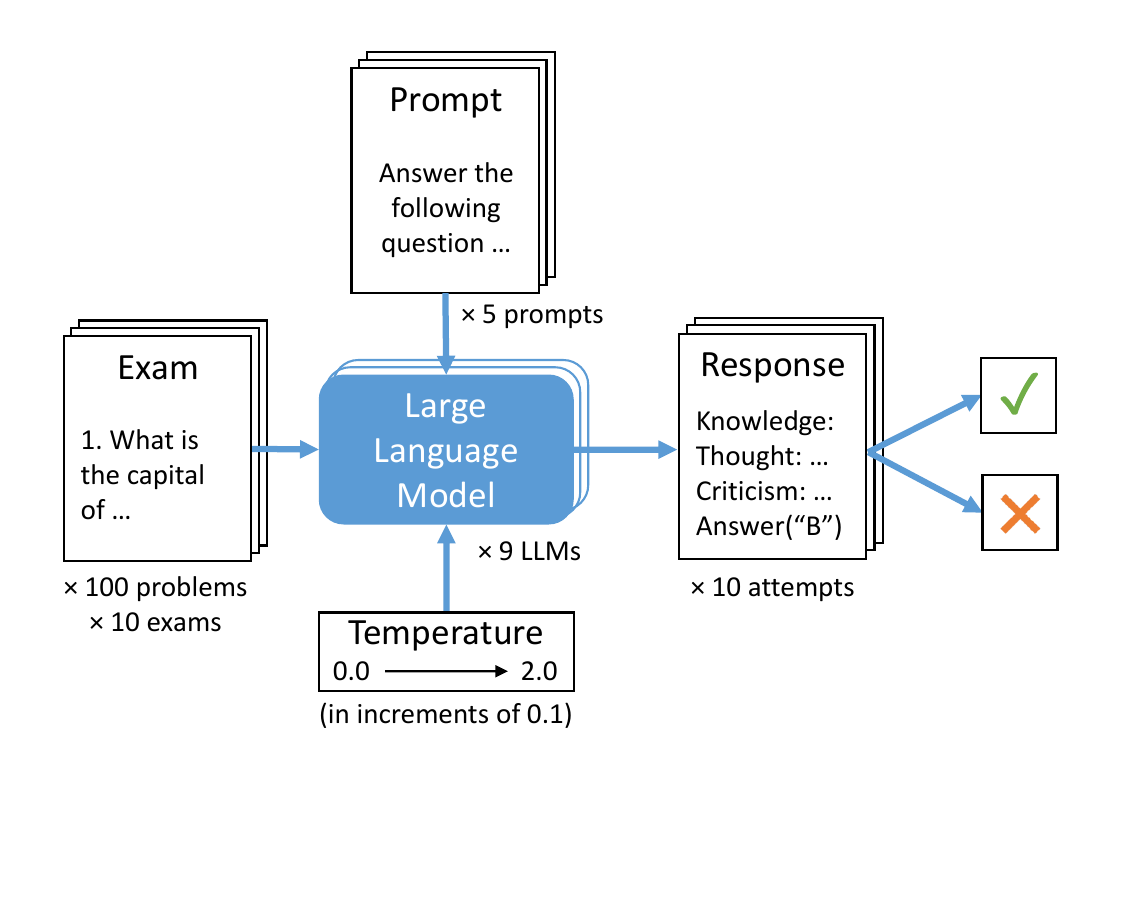}
        \caption{Diagram of the full experiment.}
        \label{fig:llm-temperature-diagram}
    \end{minipage}
    \hfill
    \begin{minipage}{0.48\textwidth}
        \begin{algorithm}[H]
            \caption{Full LLM Temperature Experiment}
            \label{alg:llm-temperature-algorithm}
            \begin{algorithmic}[1]
                \For{each model $m$ in $M$} \Comment{10 models}
                    \For{each prompt $p$ in $P$} \Comment{5 prompts}
                        \For{each exam $e$ in $E$} \Comment{10 exams}
                            \For{each temperature $\tau$ in $T$} \Comment{16 temps}
                                \For{each problem $q$ in $Q$} \Comment{100 probs}
                                    \For{each attempt $a$ in $A$} \Comment{10 attempts}
                                        \State Create the prompt
                                        \State Answer the question
                                        \State Record the answer
                                    \EndFor
                                \EndFor
                            \EndFor
                            \State Save the results
                        \EndFor
                    \EndFor
                \EndFor
                \State Process the results
                \State Analyze the results
            \end{algorithmic}
        \end{algorithm}
    \end{minipage}
\end{figure}

The full experiment setup can be seen in Figure \ref{fig:llm-temperature-diagram} and Algorithm \ref{alg:llm-temperature-algorithm}. However, due to cost and runtime considerations, we conducted a subset of the full experiment designed to capture the most valuable information as efficiently as possible.

\algrenewcommand\algorithmicindent{0.75em}

First, we instructed GPT-3.5 to complete the 100-question (small) exam using the CoT prompt with temperatures ranging from 0.0 to 2.0 in increments of 0.1. This allowed us to determine the range of viable sampling temperatures to explore. \footnote{For this experiment, we fixed all other sampling parameters (e.g., top-k, top-p, etc.) to isolate the effect of temperature.}

Performance began to drop rapidly after a temperature of 1.0 until the generated text became incoherent at 1.6. As a result, we stopped the initial temperature sweep at 1.6 and limited the rest of our sweeps from 0.0 to 1.0.

Next, we instructed the other eight LLMs to complete the 100-question (small) exam using the CoT prompt with temperatures from 0.0 to 1.0. This allowed us to determine if the results generalize to other LLMs.

Then, we instructed GPT-3.5 to complete the 100-question (small) exam using each of the five prompts over temperatures from 0.0 to 1.0. This allowed us to determine if the results generalize over various prompt-engineering techniques.

Finally, we instructed GPT-3.5 to complete the 1,000-question (large) exam using the CoT prompt with temperatures from 0.0 to 1.0. This allowed us to determine if the results were domain-specific or generalized across problem domains.

\subsection{Metrics}

To test our hypothesis, we measured the LLM’s correct-answer accuracy as our primary performance metric. For each combination of model, prompt, exam, and temperature, we calculated the accuracy as the number of correct answers from ten attempts at each problem. Then, we computed the average (mean) accuracy across all problems.

To further support our findings, we also measured the similarity of the LLM’s responses using a series of text-similarity metrics. These metrics are defined as follows:
\begin{itemize}
    \item \textbf{Jaccard similarity} – the ratio of the intersection to the union of word sets in the output text \cite{Jaccard1912}.
    \item \textbf{Bag-of-Words (BoW) similarity} – comparison of the frequency of each word, ignoring order \cite{Harris1954}.
    \item \textbf{TF-IDF similarity} – comparison of word frequency weighted by inverse document frequency \cite{SparckJones1972}.
    \item \textbf{Levenshtein similarity} – the number of edits needed to change one string of text into the other \cite{Levenshtein1966}.
    \item \textbf{BLEU score} – comparison of similarity based on n-gram overlap \cite{Papineni2001}.
    \item \textbf{SBERT similarity} – semantic similarity computed using Sentence-BERT embeddings \cite{Reimers2019}.
\end{itemize}

\subsection{Analysis}
We used the Kruskal-Wallis test to evaluate the statistical significance of any changes in accuracy as a function of temperature \cite{Kruskal1952}. We chose the Kruskal-Wallis test because the data (i.e., correct-answer accuracy by question) were not normally distributed. Rather, they were bimodally distributed with centers at 0.0 and 1.0.

\section{Results}

\subsection{Accuracy vs. Temperature}

Our analysis revealed that the problem-solving performance of LLMs remained relatively stable across sampling temperatures from 0.0 to 1.0 for all LLMs, prompt-engineering techniques, and problem domains. Using GPT-3.5 with a CoT prompt on the 1,000-question exam from 0.0 to 1.0, the Kruskal-Wallis test yielded $H(10) = 10.439, p = 0.403$.

\begin{figure}[h!]
    \centering
    \includegraphics[width=0.5\linewidth]{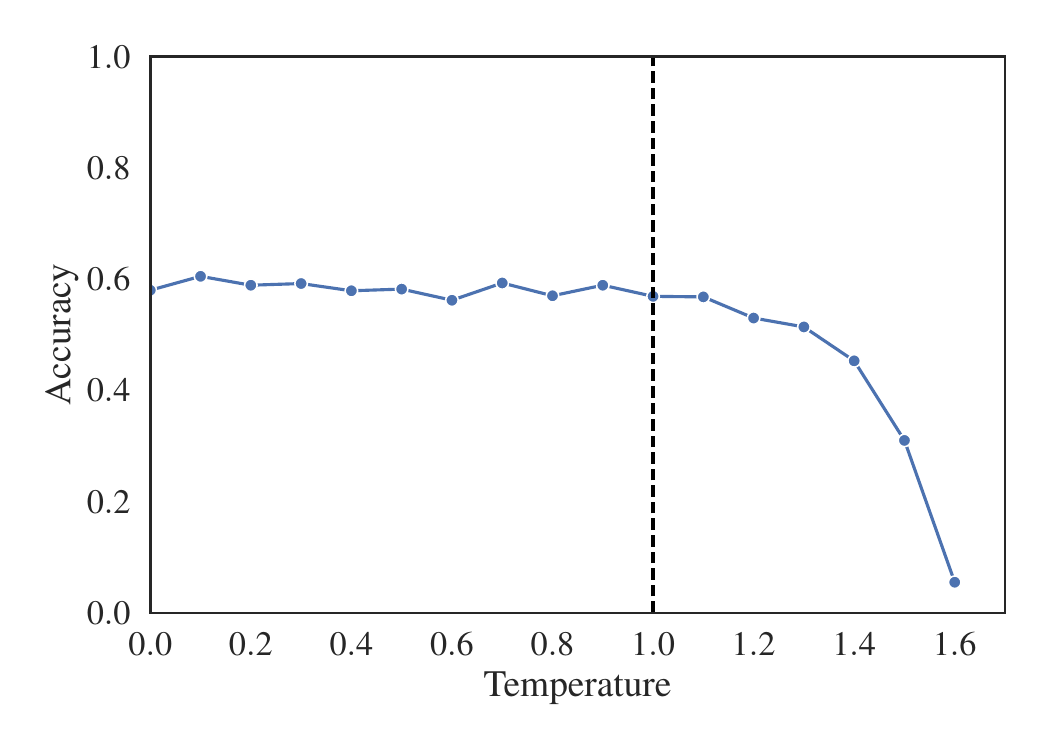}
    \caption{Accuracy by temperature from 0.0 to 1.6 for GPT-3.5 using the CoT prompt on the 100-question exam.}
    \label{fig:gpt-35-accuracy-by-temperature-extended}
\end{figure}

First, we analyzed the performance of GPT-3.5 using the CoT prompt on the 100-question exam. Accuracy remained stable over temperatures from 0.0 to 1.0. However, after a temperature of 1.0, the text rapidly became incoherent, and the accuracy began to drop until it reached zero around a temperature of 1.6. See Figure \ref{fig:gpt-35-accuracy-by-temperature-extended}.

\begin{figure}[h!]
    \centering
    \begin{minipage}{0.58\linewidth}
        \centering
        \includegraphics[width=\linewidth]{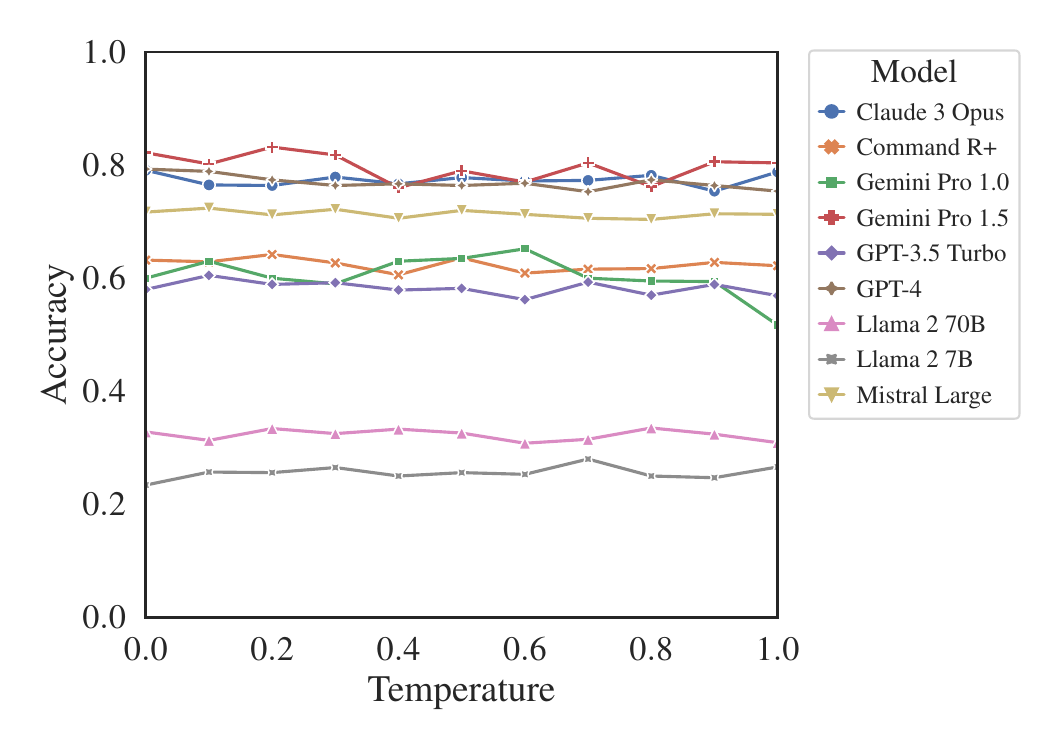}
        \captionof{figure}{Accuracy by temperature and model using the \\ CoT prompt on the 100-question exam.}
        \label{fig:accuracy-by-temperature-and-model}
    \end{minipage}
    \hfill
    \begin{minipage}{0.38\linewidth}
        \centering
        \vspace{-30 pt}
        \begin{tabular}{lrr}
            \toprule
            \textbf{Model} & \textbf{H(10)} & \textbf{p-value} \\
            \midrule
            Claude 3 Opus & 1.735 & 0.998 \\
            Command R+ & 1.771 & 0.998 \\
            Gemini Pro 1.0 & 7.379 & 0.689 \\
            Gemini Pro 1.5 & 3.119 & 0.978 \\
            GPT-3.5 Turbo & 2.042 & 0.996 \\
            GPT-4 & 3.789 & 0.956 \\
            Llama 2 70B & 3.677 & 0.961 \\
            Llama 2 7B & 17.086 & 0.072 \\
            Mistral Large & 3.069 & 0.980 \\
        \bottomrule
    \end{tabular}
        \captionof{table}{Kruskal-Wallis test results by model using the CoT prompt on the 100-question exam.}
        \label{tab:kruskal-wallis-by-model}  
    \end{minipage}
\end{figure}

Second, we analyzed the performance of all nine LLMs using the CoT prompt on the 100-question exam. Accuracy also remained stable across all of the LLMs, except for Llama 2 7B. The performance of most LLMs showed a gradual (non-significant) decrease in performance as a function of temperature. See Figure \ref{fig:accuracy-by-temperature-and-model} and Table \ref{tab:kruskal-wallis-by-model}. 

Llama 2 7B did not perform better than statistically random guesses. Its poor performance was due to generating incorrectly formatted answers (39\%) and correctly formatted but incorrect answers (36\%). Its all-or-nothing behavior at a temperature of 0.0 versus more random behavior from 0.1 to 1.0 led to a much lower, yet still non-significant, p-value.

\begin{figure}[h!]
    \centering
    \begin{minipage}{0.48\linewidth}
        \centering
        \includegraphics[width=\linewidth]{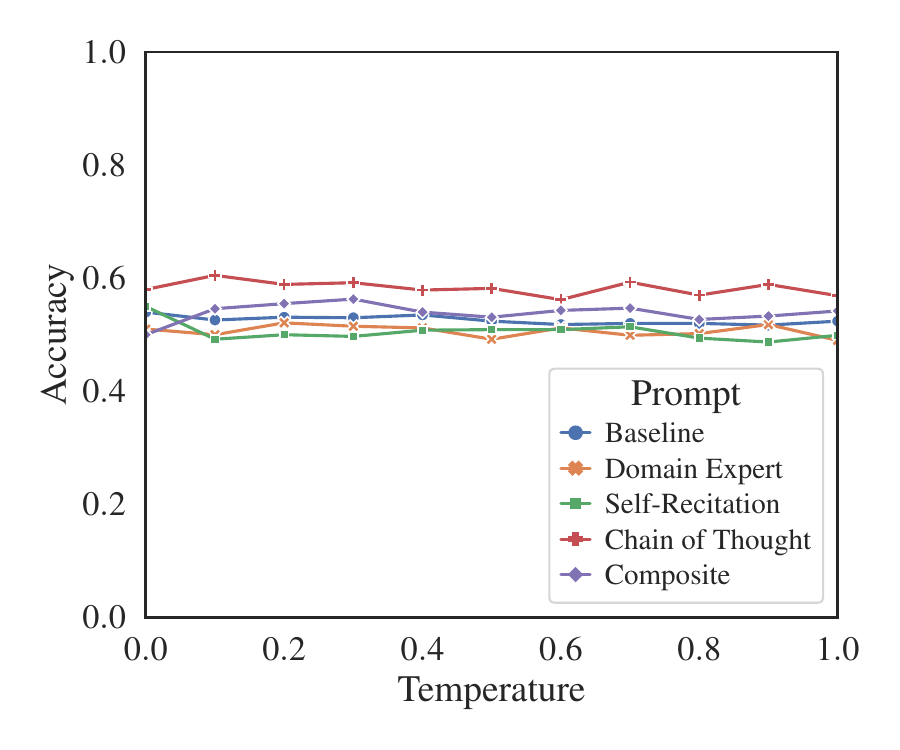}
        \captionof{figure}{Accuracy by temperature and prompt for\\ GPT-3.5 on the 100-question exam.}
        \label{fig:gpt-35-accuracy-by-temperature-and-prompt}
    \end{minipage}
    \hfill
    \begin{minipage}{0.48\linewidth}
        \centering
        \vspace{-85 pt}
        \begin{tabular}{lrr}
            \toprule
            \textbf{Prompt} & \textbf{H(10)} & \textbf{p-value} \\
            \midrule
            Baseline & 0.420 & 1.000 \\
            Domain Expert & 0.548 & 1.000 \\
            Self-recitation & 1.403 & 0.999 \\
            Chain of Thought & 2.042 & 0.996 \\
            Composite & 1.000 & 1.000 \\
            \bottomrule
        \end{tabular}
        \captionof{table}{Kruskal-Wallis test results by prompt for\\GPT-3.5 on the 100-question exam.}
        \label{tab:kruskal-wallis-by-prompt}
    \end{minipage}
\end{figure}

Third, we analyzed the performance of GPT-3.5 using each of the five prompts on the 100-question exam. Accuracy remained stable for all temperatures across all prompt-engineering techniques. The CoT prompt outperformed the other four prompts. As a result, we used the CoT prompt for all single-prompt experiments. See Figure \ref{fig:gpt-35-accuracy-by-temperature-and-prompt} and Table \ref{tab:kruskal-wallis-by-prompt}.

\begin{figure}[h]
    \centering
    \begin{minipage}{0.58\textwidth}
        \centering
        \includegraphics[width=\linewidth]{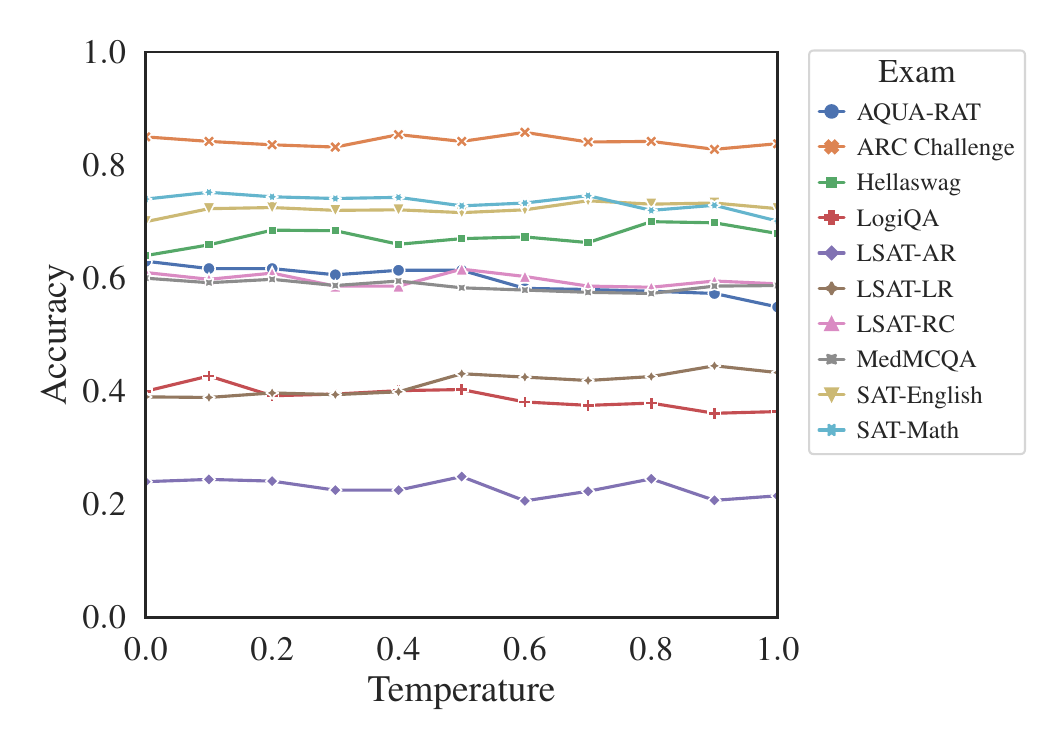}
        \captionof{figure}{Accuracy by temperature and exam for GPT-3.5\\ using the CoT prompt.}
        \label{fig:gpt-35-accuracy-by-temperature-and-exam}     
    \end{minipage}
    \hfill
    \begin{minipage}{0.38\textwidth}
        \centering
        \vspace{-40 pt}
        \begin{tabular}{lrr}
            \toprule
            \textbf{Exam} & \textbf{H(10)} & \textbf{p-value} \\
            \midrule
            AQUA-RAT & 10.320 & 0.413 \\
            ARC Challenge & 16.390 & 0.089 \\
            Hellaswag & 4.473 & 0.924 \\
            LogiQA & 3.208 & 0.976 \\
            LSAT-AR & 37.874 & < 0.001 \\
            LSAT-LR & 7.816 & 0.647 \\
            LSAT-RC & 4.037 & 0.946 \\
            MedMCQA & 2.334 & 0.993 \\
            SAT-English & 3.937 & 0.950 \\    
            SAT-Math & 21.276 & 0.019 \\
            \bottomrule
        \end{tabular}
        \captionof{table}{Kruskal-Wallis test results by exam for GPT-3.5 using the CoT prompt.}
        \label{tab:kruskal-wallis-by-exam}
    \end{minipage}
\end{figure}

Finally, we analyzed the performance of GPT-3.5 using the CoT prompt on all ten exams. Accuracy remained stable for all temperatures across all problem domains based on visual analysis. However, the LSAT-AR and SAT-Math exams showed statistically significant differences in the Kruskal-Wallis p-values. \footnote{We considered the ARC Challenge results to be non-significant since they were greater than the significance threshold of 0.05.} See Figure \ref{fig:gpt-35-accuracy-by-temperature-and-exam} and Table \ref{tab:kruskal-wallis-by-exam}.

We performed the Dunn-Bonferroni test on the LSAT-AR and SAT-Math results \cite{Dunn1964}. It revealed that the all-or-nothing behavior of responses generated at a temperature of 0.0 versus the more random responses from 0.1 to 1.0 caused the anomaly. The correct-answer accuracy for \textit{each individual problem} varied significantly when evaluated pairwise across temperatures. However, the average accuracy for \textit{all} problems remained similar across temperatures.

\subsection{Text Variability vs. Temperature}

\begin{figure}[h!]
    \centering
    \begin{minipage}{0.51\textwidth}
        \centering
        \includegraphics[width=\linewidth]{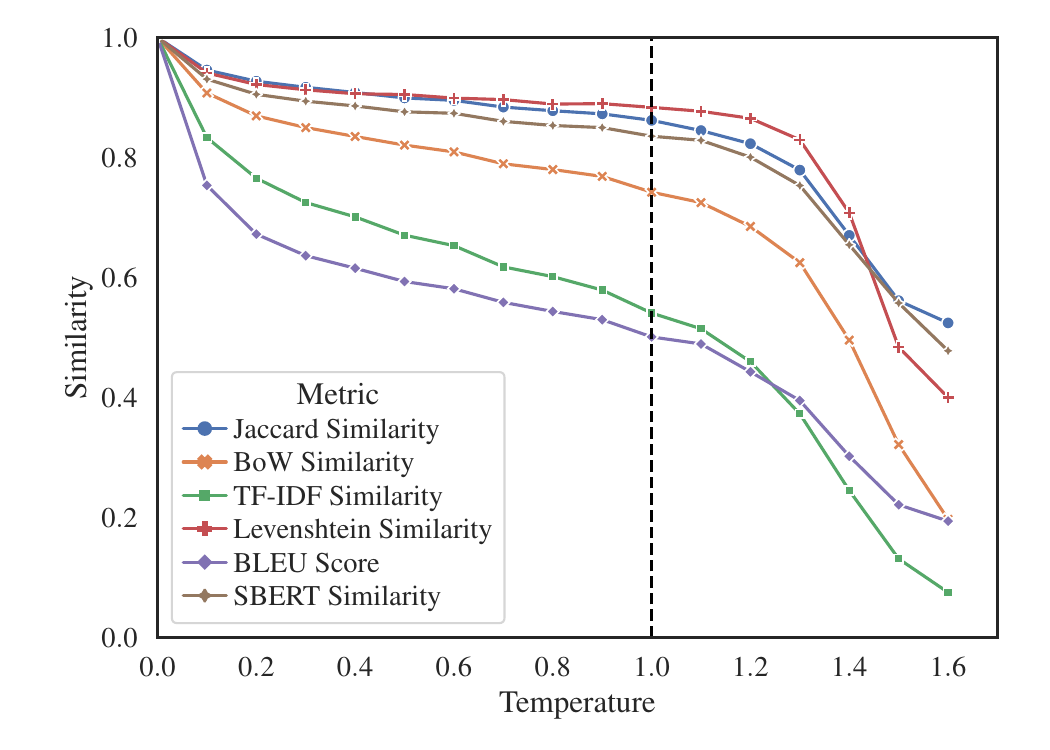}
        \caption{Text similarity by temperature and metric for GPT-3.5 using CoT prompting on the 100-question exam over sampling temperatures from 0.0 to 1.6.}
        \label{fig:text-similarity-by-temperature-and-metric}  
    \end{minipage}
    \hfill
    \begin{minipage}{0.45\textwidth}
        \centering
        \includegraphics[width=\linewidth]{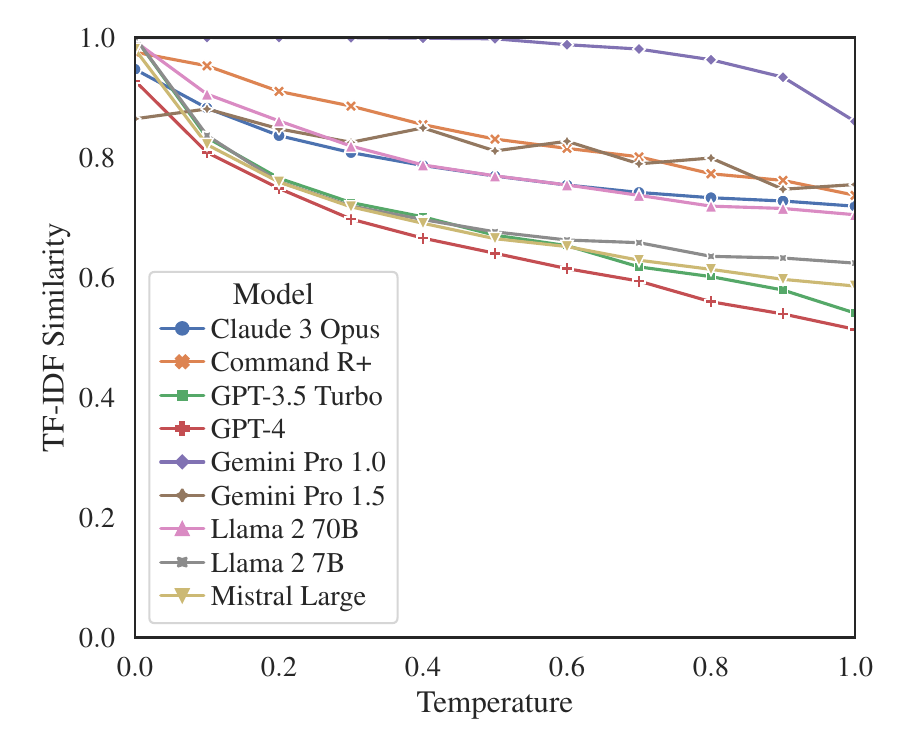}
        \caption{TF-IDF text similarity by temperature and model using the CoT prompt on the 100-question exam over sampling temperatures from 0.0 to 1.0}
        \label{fig:tf-idf-similarity-by-temperature-and-model}
    \end{minipage}
\end{figure}

To further support our results, we analyzed text variability as a function of temperature. Our findings show a clear trend of decreasing text similarity (thus increasing text variability) as temperature increases. Text similarity decreases rapidly after a temperature of 1.0, corresponding to the rapid decrease in accuracy observed above $\tau = 1.0$. See Figure \ref{fig:text-similarity-by-temperature-and-metric}.

\begin{figure}[h!]
    \centering
    \begin{minipage}{0.48\textwidth}
        \centering
        \includegraphics[width=\linewidth]{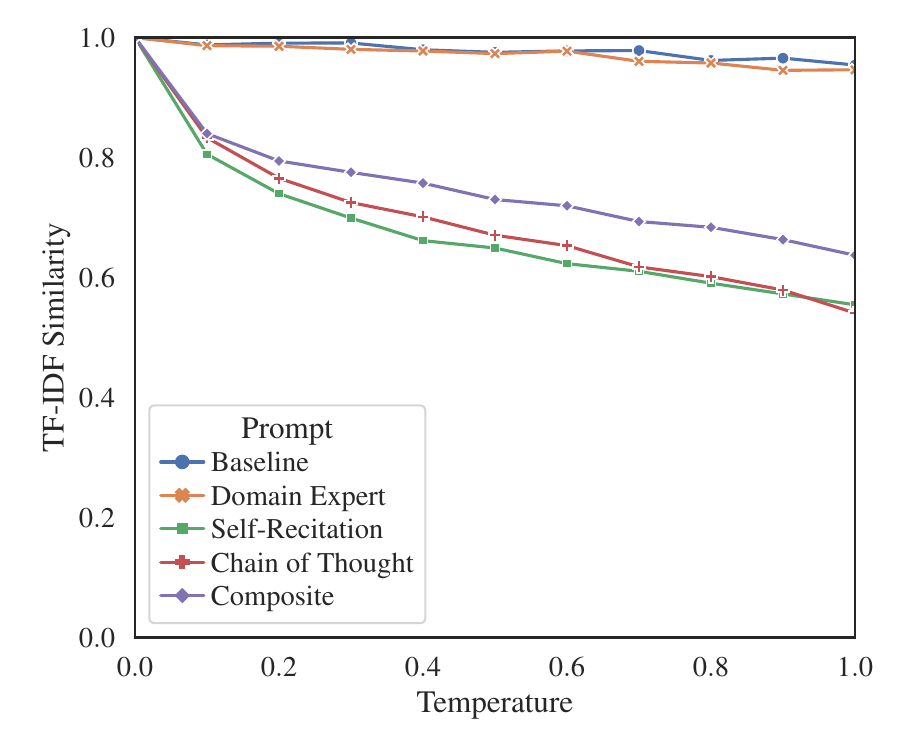}
        \caption{TF-IDF text similarity by temperature and prompt for GPT-3.5 on the 100-question exam over sampling temperatures from 0.0 to 1.0.}
        \label{fig:tf-idf-similarity-by-temperature-and-prompt}  
    \end{minipage}
    \hfill
    \begin{minipage}{0.48\textwidth}
        \centering
        \includegraphics[width=\linewidth]{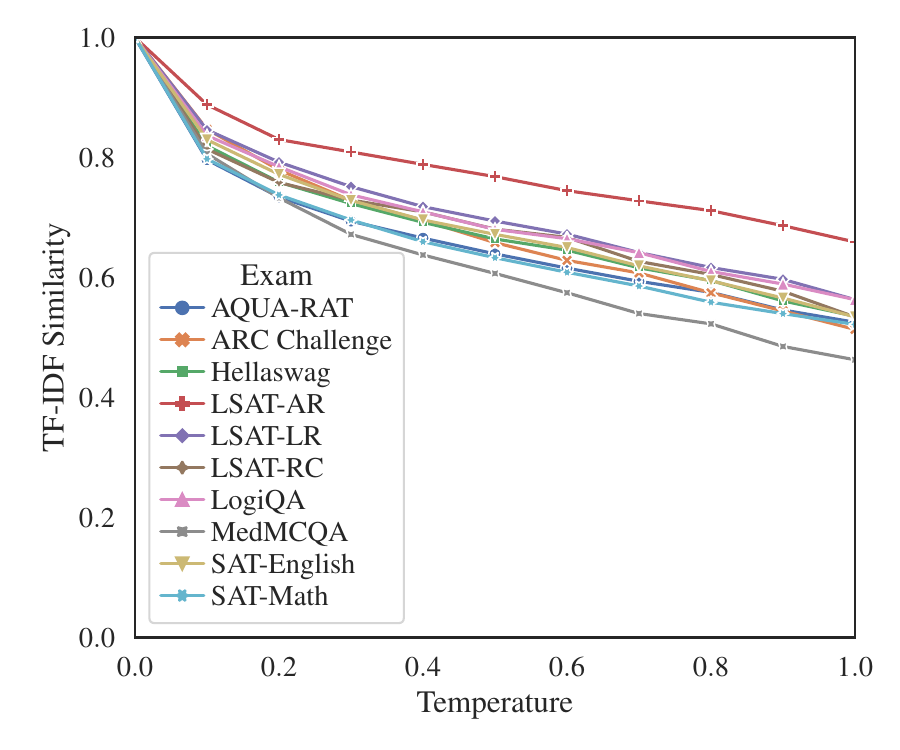}
        \caption{TF-IDF text similarity by temperature and exam for GPT-3.5 using the CoT prompt over sampling temperatures from 0.0 to 1.0}
        \label{fig:tf-idf-similarity-by-temperature-and-exam}
    \end{minipage}
\end{figure}

These results are consistent with our understanding of sampling temperature, indicating that higher temperatures produce more widely varied outputs. Furthermore, these results hold regardless of the LLM, prompt-engineering technique, or problem domain. See Figures \ref{fig:tf-idf-similarity-by-temperature-and-model}, \ref{fig:tf-idf-similarity-by-temperature-and-prompt}, and \ref{fig:tf-idf-similarity-by-temperature-and-exam}.

\section{Discussion}

\subsection{Interpretation}
Based on these results, changes in temperature from 0.0 to 1.0 do not have a statistically significant effect on the problem-solving performance of LLMs. These results appear to generalize across LLMs, prompt-engineering techniques, and problem domains. However, there are a few potential exceptions to these general findings.

Therefore, we recommend setting an LLM's sampling temperature to 0.0 for problem-solving tasks. This temperature maximizes reproducibility without compromising accuracy. In addition, it helps avoid the inevitable drop-off in performance that occurs beyond a temperature of 1.0. However, exceptions to this guidance should be taken into consideration.

\subsection{Limitations}

There were several limitations to our research study due to cost and runtime considerations:

First, our study was limited to a subset of popular LLMs. As a result, our findings may not hold for other LLMs that were excluded from our experiment.

Second, we only explored a subset of prompt-engineering techniques using a single prompt-and-response cycle with one-shot in-context learning. As a result, there may be more complex prompts or agent architectures that better leverage sampling temperature for creativity in their problem-solving capabilities.

Third, our study was limited to a subset of problems, problem domains, and problem-solving tasks. As a result, our findings may not hold for larger data sets, different problem domains, or other types of problem-solving tasks.

Fourth, due to time and cost constraints, we limited our study to two test sets of 1,000 and 100 randomly selected questions from standard benchmarks. These limited sample sizes may have introduced bias into the results. Utilizing a larger and more diverse test set would enhance the statistical reliability of our findings.

Fifth, we had to limit the sampling temperature range we explored from 0.0 to 1.0 for all combinations of models, prompts, and exams, except for GPT-3.5 using CoT prompting on the 100-question exam. As a result, the temperature hyperparameter of other LLMs may operate differently at temperatures above 1.0. 

Sixth, we fixed all other sampling parameters (e.g., top-p, top-k, repetition penalty, etc.) to isolate the effect of sampling temperature. As a result, there may be combinations of sampling parameters that result in different outcomes.

Finally, we could only explore a subset of the various combinations of models, prompts, exams, and temperatures. As a result, other combinations of LLMs, prompt-engineering techniques, and problem domains may exist where temperature plays a more important role in problem-solving performance.

\subsection{Implications}

This research study provides empirical evidence that changes in sampling temperature in the range of 0.0 to 1.0 do not significantly impact the problem-solving capabilities of LLMs on MCQA problems.

Answering this question may save AI engineers significant time and resources evaluating various sampling temperatures for their LLM agents and applications. In addition, it may reduce unproductive debates in the prompt-engineering community regarding the optimal sampling temperatures for various problem-solving tasks.

This research also provides broader insights for AI researchers studying model hallucination and problem-solution state-space search with LLMs. Our results show that increasing LLM temperature up to 1.0 does not cause the LLM to hallucinate in ways that lead to incorrect MCQA solutions. In addition, higher temperatures do not appear to improve MCQA solution-space search in ways that lead to correct solutions more often than lower temperatures.

\subsection{Future Research}

To improve upon this research, we propose the following follow-up experiments:

First, we recommend conducting this experiment with additional LLMs. Other proprietary and open-source LLMs may utilize temperature in ways that benefit their specific models but did not benefit the LLMs we tested.

Second, we recommend expanding beyond MCQA problems to other types of problem-solving tasks whose correct answers are more open-ended. The limited effects of sampling temperature in our experiments may have simply resulted from the constraints imposed by the structure of MCQA problems.

Third, we recommend conducting additional experiments with more MCQA problems and problem domains. We recommend specifically targeting tasks and problem domains that require more creative solutions or lateral “out-of-the-box” thinking.

Fourth, we recommend extending the sampling temperature range until accuracy drops to zero for each LLM, prompt, and exam. However, it should be noted that as the generated text becomes more random, the number of tokens in each response increases significantly, leading to a considerable increase in runtime and cost to explore temperatures above 1.0.

Finally, we recommend a more in-depth error analysis to determine if any sub-types of problems within these problem domains benefit from changes to sampling temperature. It is possible that statistical noise or averaging may have hidden individual problems that were sensitive to changes in sampling temperature.

\section{Conclusion}

This research study empirically investigated the effect of sampling temperature on the problem-solving performance of LLMs across multiple problem domains.

We demonstrated that changes in sampling temperature from 0.0 to 1.0 do not produce statistically significant differences in problem-solving performance on MCQA problems across multiple LLMs, prompt-engineering techniques, and problem domains.

These results have practical implications for AI engineers using LLMs to develop new AI systems. Additionally, they have theoretical implications for AI researchers studying model hallucination and solution-space search with LLMs.

\bibliographystyle{IEEEtran}
\bibliography{references}

\newpage
\appendix
\section{Appendix}

\begin{figure*}[h]
\centering
\fontsize{8pt}{8pt}\selectfont
\begin{verbatim}
[System Prompt]
You are an expert in {{expertise}}.
Your task is to answer the following multiple-choice questions.
First, you should recite all of the relevant knowledge you have about the question and each option.
Next, you should think step-by-step through the problem to ensure you have the correct answer.
Then, you should critically evaluate your thoughts to identify any flaws in your facts, logic, and reasoning.
Finally, you MUST answer the question using the following format 'Action: Answer("[choice]")'  
The parameter [choice] is the letter or number of the answer you want to select (e.g. "A", "B", "C", or "D")
For example, 'Answer("C")' will select choice "C" as the best answer.
The answer MUST ALWAYS be one of the available choices; it CANNOT be "None of the Above".
If you think the answer is "none of the above", then you MUST select the most likely answer.

[Example Problem]
Question: What is the capital of the state where Johns Hopkins University is located?
Choices:
  A: Baltimore
  B: Annapolis
  C: Des Moines
  D: Las Vegas

[Example Solution]
Knowledge: 
  Johns Hopkins University is located in Baltimore, Maryland.
  A: Baltimore is a city located in the State of Maryland, but it is not the capital of Maryland.
  B: Annapolis is the capital of the State of Maryland.
  C: Des Moines is a city located in the State of Iowa and is the capital of Iowa.
  D: Las Vegas is located in the State of Nevada, but it is not the capital of Nevada.
Thought: 
  Johns Hopkins University is located in Baltimore.
  Baltimore is a city located in the state of Maryland.
  The capital of Maryland is Baltimore.
  Therefore, the capital of the state where Johns Hopkins University is located is Baltimore.
  The answer is A: Baltimore.
Criticism: 
  You are correct that Johns Hopkins is located in the State of Maryland. 
  However, the capital of Maryland is Annapolis, not Baltimore.
  So, the correct answer is actually B: Annapolis.
Action: Answer("B")  

\end{verbatim}
\normalfont
\caption{Sample of the composite system prompt with a one-shot example (i.e., problem-and-solution pair).}
\label{fig:one-shot-example}
\end{figure*}

\begin{figure*}[h]
\centering
\fontsize{8pt}{8pt}\selectfont
\ttfamily
\begin{verbatim}
{
  "source": "arc/arc-challenge-test", 
  "source_id": 1, 
  "topic": "Science", 
  "context": "", 
  "question": "An astronomer observes that a planet rotates faster 
               after a meteorite impact. Which is the most likely effect
               of this increase in rotation?", 
  "choices": {
    "A": "Planetary density will decrease.", 
    "B": "Planetary years will become longer.", 
    "C": "Planetary days will become shorter.", 
    "D": "Planetary gravity will become stronger." }, 
  "answer": "C", 
  "solution":""
}
\end{verbatim}
\normalfont
\caption{Sample of an MCQA problem in JSON-L format – with whitespace added for readability.}
\label{fig:mcqa-sample}
\end{figure*}

\end{document}